\documentclass[]{article}
\usepackage{cmap} 
\usepackage{ifthen}
\usepackage[T1]{fontenc}
\usepackage[utf8]{inputenc}
\usepackage{amsmath}
\usepackage{booktabs}
\usepackage{color}

\usepackage[font={small,it},labelfont=bf]{caption}
\usepackage{float}

\usepackage{longtable,ltcaption,array}
\newlength{\DUtablewidth} 

\usepackage{arxiv}
\usepackage{scipy}
\makeatletter
\def\PY@reset{\let\PY@it=\relax \let\PY@bf=\relax%
    \let\PY@ul=\relax \let\PY@tc=\relax%
    \let\PY@bc=\relax \let\PY@ff=\relax}
\def\PY@tok#1{\csname PY@tok@#1\endcsname}
\def\PY@toks#1+{\ifx\relax#1\empty\else%
    \PY@tok{#1}\expandafter\PY@toks\fi}
\def\PY@do#1{\PY@bc{\PY@tc{\PY@ul{%
    \PY@it{\PY@bf{\PY@ff{#1}}}}}}}
\def\PY#1#2{\PY@reset\PY@toks#1+\relax+\PY@do{#2}}

\expandafter\def\csname PY@tok@w\endcsname{\def\PY@tc##1{\textcolor[rgb]{0.73,0.73,0.73}{##1}}}
\expandafter\def\csname PY@tok@c\endcsname{\let\PY@it=\textit\def\PY@tc##1{\textcolor[rgb]{0.25,0.50,0.56}{##1}}}
\expandafter\def\csname PY@tok@cp\endcsname{\def\PY@tc##1{\textcolor[rgb]{0.00,0.44,0.13}{##1}}}
\expandafter\def\csname PY@tok@cs\endcsname{\def\PY@tc##1{\textcolor[rgb]{0.25,0.50,0.56}{##1}}\def\PY@bc##1{\setlength{\fboxsep}{0pt}\colorbox[rgb]{1.00,0.94,0.94}{\strut ##1}}}
\expandafter\def\csname PY@tok@k\endcsname{\let\PY@bf=\textbf\def\PY@tc##1{\textcolor[rgb]{0.00,0.44,0.13}{##1}}}
\expandafter\def\csname PY@tok@kp\endcsname{\def\PY@tc##1{\textcolor[rgb]{0.00,0.44,0.13}{##1}}}
\expandafter\def\csname PY@tok@kt\endcsname{\def\PY@tc##1{\textcolor[rgb]{0.56,0.13,0.00}{##1}}}
\expandafter\def\csname PY@tok@o\endcsname{\def\PY@tc##1{\textcolor[rgb]{0.40,0.40,0.40}{##1}}}
\expandafter\def\csname PY@tok@ow\endcsname{\let\PY@bf=\textbf\def\PY@tc##1{\textcolor[rgb]{0.00,0.44,0.13}{##1}}}
\expandafter\def\csname PY@tok@nb\endcsname{\def\PY@tc##1{\textcolor[rgb]{0.00,0.44,0.13}{##1}}}
\expandafter\def\csname PY@tok@nf\endcsname{\def\PY@tc##1{\textcolor[rgb]{0.02,0.16,0.49}{##1}}}
\expandafter\def\csname PY@tok@nc\endcsname{\let\PY@bf=\textbf\def\PY@tc##1{\textcolor[rgb]{0.05,0.52,0.71}{##1}}}
\expandafter\def\csname PY@tok@nn\endcsname{\let\PY@bf=\textbf\def\PY@tc##1{\textcolor[rgb]{0.05,0.52,0.71}{##1}}}
\expandafter\def\csname PY@tok@ne\endcsname{\def\PY@tc##1{\textcolor[rgb]{0.00,0.44,0.13}{##1}}}
\expandafter\def\csname PY@tok@nv\endcsname{\def\PY@tc##1{\textcolor[rgb]{0.73,0.38,0.84}{##1}}}
\expandafter\def\csname PY@tok@no\endcsname{\def\PY@tc##1{\textcolor[rgb]{0.38,0.68,0.84}{##1}}}
\expandafter\def\csname PY@tok@nl\endcsname{\let\PY@bf=\textbf\def\PY@tc##1{\textcolor[rgb]{0.00,0.13,0.44}{##1}}}
\expandafter\def\csname PY@tok@ni\endcsname{\let\PY@bf=\textbf\def\PY@tc##1{\textcolor[rgb]{0.84,0.33,0.22}{##1}}}
\expandafter\def\csname PY@tok@na\endcsname{\def\PY@tc##1{\textcolor[rgb]{0.25,0.44,0.63}{##1}}}
\expandafter\def\csname PY@tok@nt\endcsname{\let\PY@bf=\textbf\def\PY@tc##1{\textcolor[rgb]{0.02,0.16,0.45}{##1}}}
\expandafter\def\csname PY@tok@nd\endcsname{\let\PY@bf=\textbf\def\PY@tc##1{\textcolor[rgb]{0.33,0.33,0.33}{##1}}}
\expandafter\def\csname PY@tok@s\endcsname{\def\PY@tc##1{\textcolor[rgb]{0.25,0.44,0.63}{##1}}}
\expandafter\def\csname PY@tok@sd\endcsname{\let\PY@it=\textit\def\PY@tc##1{\textcolor[rgb]{0.25,0.44,0.63}{##1}}}
\expandafter\def\csname PY@tok@si\endcsname{\let\PY@it=\textit\def\PY@tc##1{\textcolor[rgb]{0.44,0.63,0.82}{##1}}}
\expandafter\def\csname PY@tok@se\endcsname{\let\PY@bf=\textbf\def\PY@tc##1{\textcolor[rgb]{0.25,0.44,0.63}{##1}}}
\expandafter\def\csname PY@tok@sr\endcsname{\def\PY@tc##1{\textcolor[rgb]{0.14,0.33,0.53}{##1}}}
\expandafter\def\csname PY@tok@ss\endcsname{\def\PY@tc##1{\textcolor[rgb]{0.32,0.47,0.09}{##1}}}
\expandafter\def\csname PY@tok@sx\endcsname{\def\PY@tc##1{\textcolor[rgb]{0.78,0.36,0.04}{##1}}}
\expandafter\def\csname PY@tok@m\endcsname{\def\PY@tc##1{\textcolor[rgb]{0.13,0.50,0.31}{##1}}}
\expandafter\def\csname PY@tok@gh\endcsname{\let\PY@bf=\textbf\def\PY@tc##1{\textcolor[rgb]{0.00,0.00,0.50}{##1}}}
\expandafter\def\csname PY@tok@gu\endcsname{\let\PY@bf=\textbf\def\PY@tc##1{\textcolor[rgb]{0.50,0.00,0.50}{##1}}}
\expandafter\def\csname PY@tok@gd\endcsname{\def\PY@tc##1{\textcolor[rgb]{0.63,0.00,0.00}{##1}}}
\expandafter\def\csname PY@tok@gi\endcsname{\def\PY@tc##1{\textcolor[rgb]{0.00,0.63,0.00}{##1}}}
\expandafter\def\csname PY@tok@gr\endcsname{\def\PY@tc##1{\textcolor[rgb]{1.00,0.00,0.00}{##1}}}
\expandafter\def\csname PY@tok@ge\endcsname{\let\PY@it=\textit}
\expandafter\def\csname PY@tok@gs\endcsname{\let\PY@bf=\textbf}
\expandafter\def\csname PY@tok@gp\endcsname{\let\PY@bf=\textbf\def\PY@tc##1{\textcolor[rgb]{0.78,0.36,0.04}{##1}}}
\expandafter\def\csname PY@tok@go\endcsname{\def\PY@tc##1{\textcolor[rgb]{0.20,0.20,0.20}{##1}}}
\expandafter\def\csname PY@tok@gt\endcsname{\def\PY@tc##1{\textcolor[rgb]{0.00,0.27,0.87}{##1}}}
\expandafter\def\csname PY@tok@err\endcsname{\def\PY@bc##1{\setlength{\fboxsep}{0pt}\fcolorbox[rgb]{1.00,0.00,0.00}{1,1,1}{\strut ##1}}}
\expandafter\def\csname PY@tok@kc\endcsname{\let\PY@bf=\textbf\def\PY@tc##1{\textcolor[rgb]{0.00,0.44,0.13}{##1}}}
\expandafter\def\csname PY@tok@kd\endcsname{\let\PY@bf=\textbf\def\PY@tc##1{\textcolor[rgb]{0.00,0.44,0.13}{##1}}}
\expandafter\def\csname PY@tok@kn\endcsname{\let\PY@bf=\textbf\def\PY@tc##1{\textcolor[rgb]{0.00,0.44,0.13}{##1}}}
\expandafter\def\csname PY@tok@kr\endcsname{\let\PY@bf=\textbf\def\PY@tc##1{\textcolor[rgb]{0.00,0.44,0.13}{##1}}}
\expandafter\def\csname PY@tok@bp\endcsname{\def\PY@tc##1{\textcolor[rgb]{0.00,0.44,0.13}{##1}}}
\expandafter\def\csname PY@tok@fm\endcsname{\def\PY@tc##1{\textcolor[rgb]{0.02,0.16,0.49}{##1}}}
\expandafter\def\csname PY@tok@vc\endcsname{\def\PY@tc##1{\textcolor[rgb]{0.73,0.38,0.84}{##1}}}
\expandafter\def\csname PY@tok@vg\endcsname{\def\PY@tc##1{\textcolor[rgb]{0.73,0.38,0.84}{##1}}}
\expandafter\def\csname PY@tok@vi\endcsname{\def\PY@tc##1{\textcolor[rgb]{0.73,0.38,0.84}{##1}}}
\expandafter\def\csname PY@tok@vm\endcsname{\def\PY@tc##1{\textcolor[rgb]{0.73,0.38,0.84}{##1}}}
\expandafter\def\csname PY@tok@sa\endcsname{\def\PY@tc##1{\textcolor[rgb]{0.25,0.44,0.63}{##1}}}
\expandafter\def\csname PY@tok@sb\endcsname{\def\PY@tc##1{\textcolor[rgb]{0.25,0.44,0.63}{##1}}}
\expandafter\def\csname PY@tok@sc\endcsname{\def\PY@tc##1{\textcolor[rgb]{0.25,0.44,0.63}{##1}}}
\expandafter\def\csname PY@tok@dl\endcsname{\def\PY@tc##1{\textcolor[rgb]{0.25,0.44,0.63}{##1}}}
\expandafter\def\csname PY@tok@s2\endcsname{\def\PY@tc##1{\textcolor[rgb]{0.25,0.44,0.63}{##1}}}
\expandafter\def\csname PY@tok@sh\endcsname{\def\PY@tc##1{\textcolor[rgb]{0.25,0.44,0.63}{##1}}}
\expandafter\def\csname PY@tok@s1\endcsname{\def\PY@tc##1{\textcolor[rgb]{0.25,0.44,0.63}{##1}}}
\expandafter\def\csname PY@tok@mb\endcsname{\def\PY@tc##1{\textcolor[rgb]{0.13,0.50,0.31}{##1}}}
\expandafter\def\csname PY@tok@mf\endcsname{\def\PY@tc##1{\textcolor[rgb]{0.13,0.50,0.31}{##1}}}
\expandafter\def\csname PY@tok@mh\endcsname{\def\PY@tc##1{\textcolor[rgb]{0.13,0.50,0.31}{##1}}}
\expandafter\def\csname PY@tok@mi\endcsname{\def\PY@tc##1{\textcolor[rgb]{0.13,0.50,0.31}{##1}}}
\expandafter\def\csname PY@tok@il\endcsname{\def\PY@tc##1{\textcolor[rgb]{0.13,0.50,0.31}{##1}}}
\expandafter\def\csname PY@tok@mo\endcsname{\def\PY@tc##1{\textcolor[rgb]{0.13,0.50,0.31}{##1}}}
\expandafter\def\csname PY@tok@ch\endcsname{\let\PY@it=\textit\def\PY@tc##1{\textcolor[rgb]{0.25,0.50,0.56}{##1}}}
\expandafter\def\csname PY@tok@cm\endcsname{\let\PY@it=\textit\def\PY@tc##1{\textcolor[rgb]{0.25,0.50,0.56}{##1}}}
\expandafter\def\csname PY@tok@cpf\endcsname{\let\PY@it=\textit\def\PY@tc##1{\textcolor[rgb]{0.25,0.50,0.56}{##1}}}
\expandafter\def\csname PY@tok@c1\endcsname{\let\PY@it=\textit\def\PY@tc##1{\textcolor[rgb]{0.25,0.50,0.56}{##1}}}

\def\PYZus{\char`\_}

\makeatother

\providecommand*{\DUrole}[2]{%
  \ifcsname docutilsrole#1\endcsname%
    \csname docutilsrole#1\endcsname{#2}%
  \else
    \csname DUrole#1\endcsname{#2}%
  \fi%
}

\ifthenelse{\isundefined{\hypersetup}}{
  \usepackage[colorlinks=true,linkcolor=blue,urlcolor=blue]{hyperref}
  \usepackage{bookmark}
  \urlstyle{same} 
}{}

\begin{document}
\title{Quasi-orthonormal Encoding for Machine Learning Applications}\author{Haw-minn Lu\\
Gary and Mary West Health Institute\\
La Jolla, CA 92037\\
\texttt{hlu@westhealth.org}\\
}\maketitle
\InputIfFileExists{page_numbers.tex}{}{}
\newcommand*{\docutilsroleref}{\ref}
\newcommand*{\docutilsrolelabel}{\label}
\newcommand*\DUrolecode[1]{#1}
\providecommand*\DUrolecite[1]{\cite{#1}}
\begin{abstract}Most machine learning models, especially artificial neural networks, require numerical, not categorical data. We briefly describe the advantages and disadvantages of common encoding schemes. For example, one-hot encoding is commonly used for attributes with a few unrelated categories and word embeddings for attributes with many related categories (e.g., words). Neither is suitable for encoding attributes with many unrelated categories, such as diagnosis codes in healthcare applications. Application of one-hot encoding for diagnosis codes, for example, can result in extremely high dimensionality with low sample size problems or artificially induce machine learning artifacts, not to mention the explosion of computing resources needed. Quasi-orthonormal encoding (QOE) fills the gap. We briefly show how QOE compares to one-hot encoding. We provide example code of how to implement QOE using popular ML libraries such as Tensorflow and PyTorch and a demonstration of QOE to MNIST handwriting samples.\end{abstract}\keywords{machine learning, classification, categorical encoding}machine learning, classification, categorical encoding

\subsection{Introduction%
  \label{introduction}%
}

While most popular machine learning methods such as deep learning
require numerical data as input, categorical data is very common
practice. For example, a person's vitals could be a combination of both,
they could include height, weight (numerical) and gender, race
(categorical). The challenge is to convert the categorical data into a
vector of some sort.

One-hot encoding which is discussed in the next section is very
commonly used these days in machine learning but has the draw back
that it can increase the dimensionality of the data by the cardinality
of the category. For small category, this is not a significant issue
but when categories with high cardinality are present, many problems
can arise as described below.

Quasiorthonormal encoding (QOE) is a generalization of the one-hot
encoding and exploits the fact that in high dimensional vector spaces,
two random vectors are almost always orthogonal. The concept originated
with Kůrková and Kainen \DUrole{cite}{kurkova}. In many ways, QOE functions
the same as one-hot encoding but does not increase the dimensionality
of the data to the same degree as one-hot encoding. Historically, QOE
was considered for a method of encoding words but modern techinques
such as \emph{word embeddings} are now considered the state of the art
method for encoding language.

Some advantages to QOE include a reduction of dimensionality over that
of using one-hot encoding thus limiting effects of the ``curse of
dimensionality'' \DUrole{cite}{wiki:curse} or the problem of high dimension low sample size
(HDLSS). The advantage over other encodings such as binary, hash, etc.
is that it does not induce artificial geometric relationships that can
cause downstream bias in the results because each label in a category
remains mathematically near orthogonal to the other labels.

We will briefly survey \emph{classic} encoding methods, discuss the
theoretical aspects of QOE, and present a detailed example implementation
of QOE in tensorflow.

\subsection{Background%
  \label{background}%
}

Coding methods can be categorized as \emph{classic}, \emph{contrast},
\emph{Bayesian} and \emph{word embeddings}. Classic, contrast and Bayseian
encoding are given a good overview treatment by Hale's blog
\DUrole{cite}{Hale2018} with examples to be found as part of the \texttt{scikit-learn} category
encoding package \DUrole{cite}{scikit}. Both
contrast encoding and Bayesian encoding use the statistics of the data
to facilitate encoding. These two categories may be of use when more
statistical analysis is required, however there has not been widespread
adoption of these emcoding techniques for machine learning.

In a category of its own, word embeddings deserve special mention
\DUrole{cite}{wiki:wordembeddings}. Word embeddings
are used to represent words, phrases or even entire documents as a
vector. Their goal is that similar meaning or concepts get mapped to
vectors that are close in the target vector space. Additionally, it is
adapted for encoding a large categorical feature (i.e., words) into a
relatively lower dimensional space.

The remainder of the section will describe some common classic
categorical encodings

\subsubsection{Ordinal Encoding%
  \label{ordinal-encoding}%
}

To begin our overview of fundamental encoding methods, we start with
Ordinal (Label) Encoding. Ordinal encoding is the simplest and perhaps
most naive approach encoding a categorical feature -{}-{}- one simply
assigns a number to each member of a category. This is often how data
from surveys are encoded into spreadsheets for easy storage and
calculation of basic statistics. An associated data dictionary is used
to convert the values back and forth between a number and a category.
Take for example the case of gender, male could be encoded as 1 and
female as 2, with a data dictionary as follows:
:code:\texttt{\{'male': 1, 'female': 2\}}

Suppose we have three categories of ethnic groups: White, Black, and
Asian. Under ordinal encoding, suppose White is encoded as 1, Black is
encoded as 2 and Asian is encoded as 3. If a machine learning
classification is somehow confused between Asian and White and decides
to split the difference and report the in-between value (2) which
encodes Black. The issue is that arbitrary gradation between 1 and 3
introduces a natural interpolation (2) that may be nonsense. Thus, the
natural ordering of the numbers imposes an ordered geometrical
relationship between the categories that does not apply.

Nonetheless there are situations where ordinal encoding makes sense. For
example, a `rate your satisfaction' survey typically encodes five levels
(1) terrible, (2) acceptable (3) mediocre, (4) good, (5) excellent.

\subsubsection{One Hot Encoding%
  \label{one-hot-encoding}%
}

This is the most common encoding used in machine learning. One hot
encoding takes a category with cardinality $N$ and encodes each
categorical value with an $N$-dimensional vector with a single `1'
and the remainder `0's. Take as an example encoding five makes of Japanese
Cars: Toyota, Honda, Subaru, Nissan, Mitsubishi. Table \DUrole{ref}{onehot}
shows a comparison of coding between ordinal and one-hot encodings.
\begin{table}
  \begin{longtable*}{lcc}
  \toprule
  \textbf{Make} & \textbf{Ordinal} & \textbf{One-Hot} \\
  \midrule
  Toyota &  1 &  (1,0,0,0,0) \\
  Honda &  2 &  (0,1,0,0,0) \\
  Subaru &  3 &  (0,0,1,0,0) \\
  Nissan &  4 &  (0,0,0,1,0) \\
  Mitsubishi &  5 &  (0,0,0,0,1) \\
  \bottomrule
  \end{longtable*}

  \caption{Examples of Ordinal and One-Hot Encodings \DUrole{label}{onehot}}
\end{table}

The advantage is that one hot encoding doesn't induce an implicit
ordering or between categories. The primary disadvantage is that the
dimensionality of the problem has increased with corresponding increases
in complexity, computation and ``the curse of high dimensionality''.
This easily leads to the high dimensionality low sample size (HDLSS)
situation, which is a problem for most machine learning methods.

\subsubsection{Binary Encoding, Hash Encoding, BaseN Encoding%
  \label{binary-encoding-hash-encoding-basen-encoding}%
}

Somewhere in between these two are \emph{binary encoding}, \emph{hash encoding},
and \emph{baseN} encoding. Binary encoding simply labels each category with a
unique binary code and converts the binary code to a vector. Using the
previous example of the Japanese car makes, table \DUrole{ref}{binary} shows
an example of binary encoding.
\begin{table}
  \begin{longtable*}{lccc}
  \toprule
  \textbf{Make} & \textbf{Ordinal} & \textbf{as Binary} & \textbf{Binary Code} \\
  \midrule
  \endfirsthead
  Toyota &  1 &  001 &  (0,0,1) \\
  Honda &  2 &  010 &  (0,1,0) \\
  Subaru &  3 &  011 &  (0,1,1) \\
  Nissan &  4 &  100 &  (1,0,0) \\
  Mitsubishi &  5 &  101 &  (1,0,1) \\
  \bottomrule
  \end{longtable*}
  \caption{Example of Binary Codes \DUrole{label}{binary}}

\end{table}

Hash encoding assigns each category an ordinal value that is then
converted into a binary hash value that is encoded as an $n$-tuple
in the same fashion as the binary encoding. You can view hash encoding
as binary encoding applied to the hashed ordinal value. Hash encoding
has several advantages. First, it is open ended so new categories can be
added later. Second, the resultant dimensionality can be much lower than
one-hot encoding. The chief disadvantage is that categories can collide
if two categories accidentally map into the same hash value. This is a
\emph{hash collision} and must be fixed separately using a resolution
mechanism. Bernardi's blog \DUrole{cite}{hash} provides a good treatment of hash coding.

Finally, baseN encoding is a generalization of binary encoding that uses
a number base other than 2 (binary). Below is an example of the Japanese
car makes using base 3,
\begin{table}
  \begin{longtable*}{lcccc}
  \toprule
  & \textbf{as} & \textbf{Ternary} & \textbf{Balanced} \\
  \textbf{Make} & \textbf{Ordinal} & \textbf{Ternary} & \textbf{Code} & \textbf{Ternary Code} \\
  \midrule
  \endfirsthead
  Toyota & 1 & 01 & (0,1) & (0,1) \\
  Honda & 2 & 02 & (0,2) & (0,-1) \\
  Subaru & 3 & 10 & (1,0) & (1,0) \\
  Nissan & 4 & 11 & (1,1) & (1,1) \\
  Mitsubishi & 5 & 12 & (1,2) & (1,-1) \\
  \bottomrule
  \end{longtable*}
  \caption{Example of Ternary Codes}
\end{table}

A disadvantage of all three of these techniques is that while it does
reduce the dimension of the encoded feature, artificial geometric
relationships may creep in between unrelated categories. For example,
:code:\texttt{(0.7,0.7)} may be confusion between Toyota and Honda or a weak Subaru
result, although the effect is not as pronounced as ordinal encoding.

\subsubsection{Decoding%
  \label{decoding}%
}

Of course, with categorical encoding, the ability to decode an encoded vector back to a category can be very important. If the categorical variable is only an input to a machine learning system, retrieving a category may not be very important. For example, one may have a product rating model which delivers a rating based on a number of variables, some numeric like price, but others might be categorical like color, but since the output does not require category decoding, it is not important.

In an application such as categorization or imputation \DUrole{cite}{gondara}. Retrieving the category from a vector is crucial. In a training a modern classification model, a categorical output is often subject to an activation function which converts a vector into a probability of each category such as a \emph{softmax} function. Essentially, the softmax is a continuous and differential version of a ``hard max'' function which would assign a \texttt{\DUrole{code}{1}} to the vector representing the most likely category and a \texttt{\DUrole{code}{0}} to all the other categories. The conversion to a probability distribution allows the use of a negative log likelihood loss function rather than the standard root mean squared error.

Typically, other classic encoding methods use thresholds to rectify a vector first into a binary or $n$-ary value then decode the vector back to a label in accordance to the encoding. This makes them difficult to use as outputs of machine learning systems such as neural networks that rely on gradients due to lack of differentiability. Also, the decoding process is difficult to convert to a probability distribution, making negative log-likelihood or crossentropy loss functions more difficult to use.

\subsection{Theory%
  \label{theory}%
}

In this section, we'll briefly define and discuss quasiorthogonality, show how it relates to one-hot encoding and describe how this relationship can be used to develop a categorical encoding with lower cardinality.

\subsubsection{Quasiorthogonality%
  \label{quasiorthogonality}%
}

In a suitably high dimensional space, two randomly selected vectors are very likely to be nearly orthogonal or quasiorthogonal. In such an $n$-dimensional vector space, there are sets of $K$ vectors which are mutually quasiorthogonal where $K\gg n.$. A more formal definition can be stated as follows.
Given an $\epsilon$ two vectors ${\bf x}$ and
${\bf y}$ are said to be \emph{quasiorthogonal} if
$\frac{|{\bf x}\cdot {\bf y}|}{\|{\bf x}\| \|{\bf y}\|}<\epsilon$.
This extends the orthogonality principle by allowing the inner product
to not exactly equal zero. As an extension, we can define a
quasiorthonormal \emph{basis} by a set of normal vectors
$\{{\bf q}_i\}$ for $i=1,\ldots,K$ such that
$|{\bf q}_i\cdot {\bf q}_j| < \epsilon$ and
$||{\bf q}_i||=1$, for all $i,j\in\{1,\ldots,K\}$, where in
principle for large enough $n$, $K\gg n$.

The question of how large a quasiorthonormal basis can be found for a given $n$-dimensional vector space and $\epsilon$ is answered in part by the mathematical literature. \DUrole{cite}{Kainen2020} derived a lower bound for $K$ as a function of $\epsilon$
and $n$. Namely,\begin{equation*}
K \ge e^{n\epsilon^2}.
\end{equation*}This means that given an $\epsilon$, the size of potential quasiorthonormal basis grows at least exponentially as $n$ grows.

\subsubsection{One Hot Encoding Revisited%
  \label{one-hot-encoding-revisited}%
}

The method to exploit quasiorthogonality in categorical encoding
parallels the use of orthonormal basis in one-hot encoding. In machine
learning, the typical aspects of one hot encoding maps a
variable with $n$ categories into a set of unit vectors in a
$n$-dimensional space: $L=\{l_i\}$ for $i=1\ldots n$,
then the one hot encoding $E_L:L \mapsto \mathbb{R}^n$
given by $l_i \mapsto \mathbf{u}_i$ where $\mathbf{u}_i$ is
an orthonormal basis in $\mathbb{R}^n$. The simplest basis used is
$\mathbf{u}_i = (0,0,\ldots, 1, 0,\ldots, 0)$ where the $1$
is in the $i$th position which is know as the \emph{standard basis}
for $\mathbb{R}^n$.

Mapping of a vector back to the original category uses the \emph{argmax}
function, so for a vector $\mathbf{z}$,
$\mathrm{argmax}(\mathbf{z}) = i$ where $z_i>z_j$ for all
$j\ne i$ and the vector $\mathbf{z}$ decodes to
$l_{\mathrm{argmax}(\mathbf{z})}$. Of course, the argmax function
is not easily differentiable which presents problems in ML learning algorithms
that require derivatives. To fix this, a \emph{softer} version is used called
the \emph{softargmax} or now as simply \emph{softmax} and is defined as follows:\begin{equation}
\label{eq:csoftmax}
\mathrm{softmax}(\mathbf{z})_i=\frac{e^{z_i}}{\sum_{j=1}^n e^{z_j}}
\end{equation}for $i=1,2,\ldots,n$ and
${\bf z}=(z_1, z_2,\ldots, z_n) \in \mathbb{R}^n$ where
$\mathbf{z}$ is the vector being decoded. The softmax function
decodes a one-hot encoded vector into a probability density function
which enables application of negative log likelihood loss functions or
cross entropy losses.

Though one-hot encoding uses unit vectors
with one \texttt{\DUrole{code}{1}} in the vector hence a \emph{hot} component. The
formalization of the one hot encoding above allows \emph{any} orthonormal
basis to be used. So to use a generalized one-hot encoding with
orthonormal basis ${\mathbf{u}_i}$, one would map the label
$j$ to ${\mathbf{u}_j}$ for encoding where the
${\mathbf{u}_i}$ no longer have to take the standard basis form.
To decode an encoded value in this framework, we would take\begin{equation}
\label{eq:argmax}
i = \mathrm{argmax}(\mathbf{z}\cdot\mathbf{u}_1,\mathbf{z}\cdot\mathbf{u}_2,\ldots,\mathbf{z}\cdot\mathbf{u}_n).
\end{equation}This reduces to $\mathrm{argmax}(\mathbf{z})$ for the standard
basis. Thus, the softmax function can be expressed as the following,\begin{equation}
\label{eq:gsoftmax}
\mathrm{softmax}({\bf z})_i={e^{{\bf z}\cdot {\bf u}_i}\over \sum_{j=1}^n e^{{\bf z}\cdot {\bf u}_j}}.
\end{equation}

\subsubsection{Encoding%
  \label{encoding}%
}
The principle behind QOE is simple. A quasiorthonormal basis $\{{\bf q}_i\}$ is
substituted for the orthonormal basis $\{{\bf u}_i\}$ described above. So given a
quasiorthonormal basis, we can define a QOE for a set $L=\{l_i\}$
by $l_i \mapsto \mathbf{q}_i$.

Decoding $\mathbf{z}$ under QOE would use a \emph{qargmax} function analogous to the argmax function for one-hot encoding as shown in equation \DUrole{ref}{eq:qargmax} which is nearly identical to equation \DUrole{ref}{eq:argmax}.\begin{equation}
\label{eq:qargmax}
i = \mathrm{argmax}(\mathbf{z}\cdot\mathbf{q}_1,\mathbf{z}\cdot\mathbf{q}_2,\ldots,\mathbf{z}\cdot\mathbf{q}_n)
\end{equation}Analogous to the softmax function shown of equation \DUrole{ref}{eq:gsoftmax}, is a \emph{qsoftmax} function which can be expressed as\begin{equation}
\label{eq:qsoftmax}
\mathrm{qsoftmax}({\bf z})_i={e^{{\bf z}\cdot {\bf q}_i}\over \sum_{j=1}^K
e^{{\bf z}\cdot {\bf q}_j}}
\end{equation}The only real difference in the formulation is that while still
operating in ${\mathbb R}^n$ we are encoding $K>n$ labels.

Returning to our example of Japanese car makes, Table \DUrole{ref}{qoe} shows one-hot encoding and QOE of the five manufacturers. In the table, encodings are represented simply as vectors where $\mathbf{u}_i$ are unit vectors in $\mathbb{R}^5$ and
${\mathbf{q}_i}$ are a set of quasiorthonormal vectors in $\mathbb{R}^3$. It can be shown that such a quasiorthonormal can be found in \DUrole{cite}{sphere} with the minimum mutual angle of 66$^\circ$. In short, the difference between one-hot encoding and QOE is that the one-hot requires 5 dimensions and in this case QOE requires only 3.
\begin{table}
  \begin{longtable*}{lccc}
  \toprule
  \textbf{make} & \textbf{Ordinal} & \textbf{One-Hot} & \textbf{QOE} \\
  \midrule
  \endfirsthead
  Toyota & 1 & $\mathbf{u}_1$ & $\mathbf{q}_1$ \\
  Honda & 2 & $\mathbf{u}_2$ & $\mathbf{q}_2$ \\
  Subaru & 3 & $\mathbf{u}_3$ & $\mathbf{q}_3$ \\
  Nissan & 4 & $\mathbf{u}_4$ & $\mathbf{q}_4$ \\
  Mitsubishi & 5 & $\mathbf{u}_5$ & $\mathbf{q}_5$ \\
  \bottomrule
  \end{longtable*}
  \caption{Example of Quasiorthonormal Encoding \DUrole{label}{qoe}}

\end{table}

\subsection{Implementation%
  \label{implementation}%
}

\subsubsection{Mathematical%
  \label{mathematical}%
}

While equations \DUrole{ref}{eq:qargmax} and \DUrole{ref}{eq:qsoftmax} describe precisely mathematically how to implement decoding and activation functions. Literal implementation would not exploit the modern vectorized and accelerated computation available in such packages as \texttt{numpy}, \texttt{tensorflow} and \texttt{pytorch}.

To better exploit built-in functions of these packages, we define the following $n\times K$ \emph{change of coordinates} matrix\begin{equation*}
\mathbf{Q}=  \left[\begin{matrix}
\bigg| & \bigg| & &\bigg | \\
\mathbf{q}_1 & \mathbf{q}_2 & \cdots & \mathbf{q}_K \\
\bigg| & \bigg| & &\bigg | \end{matrix}\right].
\end{equation*}that transforms between the QOE space and the one hot encoding space. So
given a argmax or softmax function, we can express the quasiorthonormal
variant as follows\begin{equation*}
\mathrm{qargmax}(\mathbf{z}) = \mathrm{argmax}(\mathbf{Qz})
\end{equation*}and\begin{equation}
\label{eq:convert}
\mathrm{qsoftmax}(\mathbf{z}) = \mathrm{softmax}(\mathbf{Qz}).
\end{equation}This facilitates the use of optimized functions such as \texttt{softmax} in libraries
like \texttt{tensorflow} and using the above matrix enables quick
implementation of QOE into these packages. Not only will using native functions accelerated performance it can exploit features such as auto differentiation built into the native functions. A useful property if one wishes to use the qsoftmax function as an activation function.

Since the matrix manipulation operations and input/output shape definitions differ from package to package, we provide a qsoftmax implementation in several popular packages. In order to facilitate the most general format possible, in our examples, we will express the quasiorthogonal basis as an list of list, but the input and the output is expressed in the appropriate native class (e.g. \texttt{\DUrole{code}{numpy.ndarray}} in \texttt{numpy}).

\subsubsection{Numpy%
  \label{numpy}%
}

For Numpy, the implementation is straight-forward and follows equation \DUrole{ref}{eq:convert} almost literally and is given below.\vspace{1mm}
\begin{Verbatim}[commandchars=\\\{\},fontsize=\footnotesize]
\PY{k}{def} \PY{n+nf}{qsoftmax}\PY{p}{(}\PY{n}{x}\PY{p}{,} \PY{n}{basis}\PY{p}{)}\PY{p}{:}
    \PY{n}{qx} \PY{o}{=} \PY{n}{np}\PY{o}{.}\PY{n}{matmul}\PY{p}{(}\PY{n}{np}\PY{o}{.}\PY{n}{asarray}\PY{p}{(}\PY{n}{basis}\PY{p}{)}\PY{p}{,}\PY{n}{x}\PY{p}{)}
    \PY{k}{return} \PY{n}{softmax}\PY{p}{(}\PY{n}{qx}\PY{p}{)}
\end{Verbatim}
\vspace{1mm}
This can be wrapped in a function factory or metafunction in applications where an unparameterized activation function is required. This metafunction returns a function \texttt{qsoftmax} function for a given basis.\vspace{1mm}
\begin{Verbatim}[commandchars=\\\{\},fontsize=\footnotesize]
\PY{k}{def} \PY{n+nf}{qsoftmax}\PY{p}{(}\PY{n}{basis}\PY{p}{)}\PY{p}{:}
    \PY{k}{def} \PY{n+nf}{func}\PY{p}{(}\PY{n}{x}\PY{p}{)}\PY{p}{:}
        \PY{n}{qx} \PY{o}{=} \PY{n}{np}\PY{o}{.}\PY{n}{matmul}\PY{p}{(}\PY{n}{np}\PY{o}{.}\PY{n}{asarray}\PY{p}{(}\PY{n}{basis}\PY{p}{)}\PY{p}{,}\PY{n}{x}\PY{p}{)}
        \PY{k}{return} \PY{n}{softmax}\PY{p}{(}\PY{n}{qx}\PY{p}{)}
    \PY{k}{return} \PY{n}{func}
\end{Verbatim}
\vspace{1mm}
The \texttt{softmax} can be found in \texttt{scipy.special.softmax} or can easily be written as\vspace{1mm}
\begin{Verbatim}[commandchars=\\\{\},fontsize=\footnotesize]
\PY{k}{def} \PY{n+nf}{softmax}\PY{p}{(}\PY{n}{x}\PY{p}{)}\PY{p}{:}
     \PY{n}{ex}\PY{o}{=}\PY{n}{np}\PY{o}{.}\PY{n}{exp}\PY{p}{(}\PY{n}{x}\PY{p}{)}
     \PY{k}{return} \PY{n}{ex}\PY{o}{/}\PY{n}{np}\PY{o}{.}\PY{n}{sum}\PY{p}{(}\PY{n}{ex}\PY{p}{)}
\end{Verbatim}
\vspace{1mm}

\subsubsection{Tensorflow%
  \label{tensorflow}%
}
The following segment of code is an implementation of the \texttt{qsoftmax} using \texttt{tensorflow} functions. By using native \texttt{tensorflow} functions, the resultant \texttt{qsoftmax} function will be automatically differentiated in a backwards neural network pass.\vspace{1mm}
\begin{Verbatim}[commandchars=\\\{\},fontsize=\footnotesize]
\PY{k}{def} \PY{n+nf}{qsoftmax}\PY{p}{(}\PY{n}{x}\PY{p}{,} \PY{n}{basis}\PY{p}{)}\PY{p}{:}
    \PY{n}{qx} \PY{o}{=} \PY{n}{tf}\PY{o}{.}\PY{n}{matmul}\PY{p}{(}\PY{n}{tf}\PY{o}{.}\PY{n}{constant}\PY{p}{(}\PY{n}{basis}\PY{p}{)}\PY{p}{,} \PY{n}{x}\PY{p}{,}
                   \PY{n}{transpose\PYZus{}b}\PY{o}{=}\PY{k+kc}{True}\PY{p}{)}
    \PY{k}{return} \PY{n}{tf}\PY{o}{.}\PY{n}{nn}\PY{o}{.}\PY{n}{softmax}\PY{p}{(}\PY{n}{tf}\PY{o}{.}\PY{n}{transpose}\PY{p}{(}\PY{n}{qx}\PY{p}{)}\PY{p}{)}
\end{Verbatim}
\vspace{1mm}
The wrapped metafunction version of \texttt{qsoftmax} is also presented as this will be used below in our example of MNIST handwriting classification employing QOE.\vspace{1mm}
\begin{Verbatim}[commandchars=\\\{\},fontsize=\footnotesize]
\PY{k}{def} \PY{n+nf}{qsoftmax}\PY{p}{(}\PY{n}{basis}\PY{p}{)}\PY{p}{:}
    \PY{k}{def} \PY{n+nf}{func}\PY{p}{(}\PY{n}{x}\PY{p}{)}\PY{p}{:}
        \PY{n}{qx} \PY{o}{=} \PY{n}{tf}\PY{o}{.}\PY{n}{matmul}\PY{p}{(}\PY{n}{tf}\PY{o}{.}\PY{n}{constant}\PY{p}{(}\PY{n}{basis}\PY{p}{)}\PY{p}{,} \PY{n}{x}\PY{p}{,}
                       \PY{n}{transpose\PYZus{}b}\PY{o}{=}\PY{k+kc}{True}\PY{p}{)}
        \PY{k}{return} \PY{n}{tf}\PY{o}{.}\PY{n}{nn}\PY{o}{.}\PY{n}{softmax}\PY{p}{(}\PY{n}{tf}\PY{o}{.}\PY{n}{transpose}\PY{p}{(}\PY{n}{qx}\PY{p}{)}\PY{p}{)}
    \PY{k}{return} \PY{n}{func}
\end{Verbatim}
\vspace{1mm}

\subsubsection{Pytorch%
  \label{pytorch}%
}
Presented below is a version of the \texttt{qsoftmax} function implemented using \texttt{pytorch} primitives. The use of the \texttt{squeeze} and \texttt{unsqueeze} operations convert between a 1-dimensional vector and a 2-dimension matrix having one column. This function is only designed to accept vector inputs. In some models, especially image related models, outputs of some layers maybe multidimensional arrays. If your use case requires a multidimensional imput to the \texttt{qsoftmax} function the code may need alteration.\vspace{1mm}
\begin{Verbatim}[commandchars=\\\{\},fontsize=\footnotesize]
\PY{k}{def} \PY{n+nf}{qsoftmax}\PY{p}{(}\PY{n}{x}\PY{p}{,} \PY{n}{basis}\PY{p}{)}\PY{p}{:}
   \PY{n}{qx} \PY{o}{=} \PY{n}{torch}\PY{o}{.}\PY{n}{mm}\PY{p}{(}\PY{n}{torch}\PY{o}{.}\PY{n}{tensor}\PY{p}{(}\PY{n}{basis}\PY{p}{)}\PY{p}{,}
                 \PY{n}{x}\PY{o}{.}\PY{n}{unsqueeze}\PY{p}{(}\PY{l+m+mi}{0}\PY{p}{)}\PY{o}{.}\PY{n}{t}\PY{p}{(}\PY{p}{)}\PY{p}{)}\PY{o}{.}\PY{n}{t}\PY{p}{(}\PY{p}{)}\PY{o}{.}\PY{n}{squeeze}\PY{p}{(}\PY{p}{)}
   \PY{k}{return} \PY{n}{torch}\PY{o}{.}\PY{n}{nn}\PY{o}{.}\PY{n}{functional}\PY{o}{.}\PY{n}{softmax}\PY{p}{(}\PY{n}{qx}\PY{p}{,}\PY{n}{dim}\PY{o}{=}\PY{l+m+mi}{0}\PY{p}{)}
\end{Verbatim}
\vspace{1mm}

\subsection{Construction of an Quasiorthonormal set%
  \label{construction-of-an-quasiorthonormal-set}%
}
It is difficult find explicit constructions of quasiorthonormal sets in
the literature. Several methods are mentioned by Kainen \DUrole{cite}{kainan}, but
these constructions are theoretical and hard
to follow. There are a number of combinatorial problems related such as
spherical codes \DUrole{cite}{wiki:spheres} and Steiner Triple Systems \DUrole{cite}{wiki:steiner}, which strive to find optimal solutions. These are extremely complicated mathematical constructions and not every optimal solution has been found.

As a practical matter, optimal solutions are not necessary as long as the desired characteristics of the quasiorthonormal basis are obtained. As an example, while an optimal solution finds 28 quasiorthonormal vectors with dot products of 0.5 or under are possible in seven dimensions, you may only need 10 vectors. In other words, a suboptimal solution may yield fewer vectors than are possible for a given dimension, or a larger dimension may be required to obtain the desired number of vectors than is theoretically needed.

One practical way to construct a quasiorthonormal basis is to use spherical codes which has been studied in greater detail. Spherical codes try to find a set of points on the $n$-dimensional hypersphere
such that the minimum distance between two points is maximized. In most
constructions of spherical codes, a given point's antipodal point is
also in that code set. So in order to get a quasiorthogonal set, for
each pair of antipodal points, only one element of the pair is selected. Perhaps to better understand the relationship, between quasiorthonormal basis and spherical codes is that a set of spherical codes can be constructed by taking every vector in a quasiorthonormal basis and add its antipodal point.

The area of algorithmically finding a quasiorthonormal basis is scant as is in the related area of finding suboptimal spherical codes. However, one such method was investigated by Gautam and Vaintrob \DUrole{cite}{Gautam2013ANA}. Perhaps the easiest way to obtain a quasiorthonormal basis is to use spherical codes as described above but obtain the spherical code from the vast compliation of sphere codes by Sloane \DUrole{cite}{sphere}.

\subsection{Simple Example and Comparison%
  \label{simple-example-and-comparison}%
}

To demonstrate how QOE can be used in machine learning, we provide a simple experiment/demonstration.
This demonstration in addition to showing how to construct a classification system using QOE gives an sense of the effect of QOE on accuracy. As an initial experiment, we applied QOE to classification of the Modified National Institute of Standards and Technology (MNIST) handwriting dataset \DUrole{cite}{mnist}, using the 60000 training examples with 10000 test
examples. As there are 10 categories, we needed sets of quasiorthonormal
bases with 10 elements. We took the spherical code for 24 points in
4-dimensions, giving us 12 quasi-orthogonal vectors. The maximum
pairwise dot product was 0.5 leading to an angle of 60$^\circ$.
We also took the spherical code for 56 points in 7-dimensions, giving 28
quasi-orthogonal vectors. The maximum pairwise dot product was .33
leading to an angle of a little over 70$^\circ$

We used a hidden layer with 64 units with a ReLU activation function.
Next there is a 20\% dropout layer to mitigate overtraining, then an
output layer whose width depends on the encoding used. We elected for this demonstration to use one of the simplest models hence there are no convolutional or pooling layers used as often seen in other sample MNIST handwriting classifers. The following example is implemented using \texttt{tensorflow} and \texttt{keras}.

\subsubsection{Validating the QSoftmax Function%
  \label{validating-the-qsoftmax-function}%
}

We begin by validating the \texttt{qsoftmax} function as provided above. This is done by first constructing a reference model built on \texttt{tensorflow} and \texttt{keras} in the standard way. In fact this example is nearly identical to the presented in the \emph{Quickstart for Beginners} guide \DUrole{cite}{tensorflow} for \texttt{tensorflow} with the exception that we employ a separate \texttt{Activation} for clarity.\vspace{1mm}
\begin{Verbatim}[commandchars=\\\{\},fontsize=\footnotesize]
\PY{n}{normal\PYZus{}model} \PY{o}{=} \PY{n}{tf}\PY{o}{.}\PY{n}{keras}\PY{o}{.}\PY{n}{models}\PY{o}{.}\PY{n}{Sequential}\PY{p}{(}\PY{p}{[}
  \PY{n}{tf}\PY{o}{.}\PY{n}{keras}\PY{o}{.}\PY{n}{layers}\PY{o}{.}\PY{n}{Flatten}\PY{p}{(}\PY{n}{input\PYZus{}shape}\PY{o}{=}\PY{p}{(}\PY{l+m+mi}{28}\PY{p}{,} \PY{l+m+mi}{28}\PY{p}{)}\PY{p}{)}\PY{p}{,}
  \PY{n}{tf}\PY{o}{.}\PY{n}{keras}\PY{o}{.}\PY{n}{layers}\PY{o}{.}\PY{n}{Dense}\PY{p}{(}\PY{l+m+mi}{64}\PY{p}{,} \PY{n}{activation}\PY{o}{=}\PY{n}{tf}\PY{o}{.}\PY{n}{nn}\PY{o}{.}\PY{n}{relu}\PY{p}{)}\PY{p}{,}
  \PY{n}{tf}\PY{o}{.}\PY{n}{keras}\PY{o}{.}\PY{n}{layers}\PY{o}{.}\PY{n}{Dropout}\PY{p}{(}\PY{l+m+mf}{0.2}\PY{p}{)}\PY{p}{,}
  \PY{n}{tf}\PY{o}{.}\PY{n}{keras}\PY{o}{.}\PY{n}{layers}\PY{o}{.}\PY{n}{Dense}\PY{p}{(}\PY{l+m+mi}{10}\PY{p}{)}
  \PY{n}{tf}\PY{o}{.}\PY{n}{keras}\PY{o}{.}\PY{n}{layers}\PY{o}{.}\PY{n}{Activation}\PY{p}{(}\PY{n}{tf}\PY{o}{.}\PY{n}{nn}\PY{o}{.}\PY{n}{softmax}\PY{p}{)}
\PY{p}{]}\PY{p}{)}
\end{Verbatim}
\vspace{1mm}
To validate that the \texttt{qsoftmax} function and the use of a \texttt{Lambda} layer is propery used, the \texttt{qsoftmax} metafunction is used with the identity matrix to represent the basis. Mathematically, the resultant \texttt{qsoftmax} function in the \texttt{Lambda} layer is exactly the \texttt{softmax} function.  The code is shown below:\vspace{1mm}
\begin{Verbatim}[commandchars=\\\{\},fontsize=\footnotesize]
\PY{n}{sanity\PYZus{}model} \PY{o}{=} \PY{n}{tf}\PY{o}{.}\PY{n}{keras}\PY{o}{.}\PY{n}{models}\PY{o}{.}\PY{n}{Sequential}\PY{p}{(}\PY{p}{[}
  \PY{n}{tf}\PY{o}{.}\PY{n}{keras}\PY{o}{.}\PY{n}{layers}\PY{o}{.}\PY{n}{Flatten}\PY{p}{(}\PY{n}{input\PYZus{}shape}\PY{o}{=}\PY{p}{(}\PY{l+m+mi}{28}\PY{p}{,} \PY{l+m+mi}{28}\PY{p}{)}\PY{p}{)}\PY{p}{,}
  \PY{n}{tf}\PY{o}{.}\PY{n}{keras}\PY{o}{.}\PY{n}{layers}\PY{o}{.}\PY{n}{Dense}\PY{p}{(}\PY{l+m+mi}{64}\PY{p}{,} \PY{n}{activation}\PY{o}{=}\PY{n}{tf}\PY{o}{.}\PY{n}{nn}\PY{o}{.}\PY{n}{relu}\PY{p}{)}\PY{p}{,}
  \PY{n}{tf}\PY{o}{.}\PY{n}{keras}\PY{o}{.}\PY{n}{layers}\PY{o}{.}\PY{n}{Dropout}\PY{p}{(}\PY{l+m+mf}{0.2}\PY{p}{)}\PY{p}{,}
  \PY{n}{tf}\PY{o}{.}\PY{n}{keras}\PY{o}{.}\PY{n}{layers}\PY{o}{.}\PY{n}{Dense}\PY{p}{(}\PY{l+m+mi}{10}\PY{p}{)}
  \PY{n}{tf}\PY{o}{.}\PY{n}{keras}\PY{o}{.}\PY{n}{layers}\PY{o}{.}\PY{n}{Lambda}\PY{p}{(}\PY{n}{qsoftmax}\PY{p}{(}\PY{n}{numpy}\PY{o}{.}\PY{n}{identity}\PY{p}{(}\PY{l+m+mi}{10}\PY{p}{,}
                             \PY{n}{dtype}\PY{o}{=}\PY{n}{numpy}\PY{o}{.}\PY{n}{float32}\PY{p}{)}\PY{p}{)}\PY{p}{)}
\PY{p}{]}\PY{p}{)}
\end{Verbatim}
\vspace{1mm}
This should function identically as the reference model because it tests
that the qsoftmax function operates as expected (which it does in this case).
This is useful for troubleshooting if you have difficulty.

\subsubsection{Examples on Quasiorthogonal Basis%
  \label{examples-on-quasiorthogonal-basis}%
}

To recap, for the two QOE experiments we take a set of 10 mutually quasiorthonormal vectors from a four dimensional space and from a seven dimensional space all derived from spherical codes from tables mentioned above and only took 10 vectors. For the code, the basis for each experiment are labeled  \texttt{basis4} and \texttt{basis7}, respectively. This leads to the following models, \texttt{basis4\_model} and \texttt{basis7\_model}.\vspace{1mm}
\begin{Verbatim}[commandchars=\\\{\},fontsize=\footnotesize]
\PY{n}{basis4\PYZus{}model} \PY{o}{=} \PY{n}{tf}\PY{o}{.}\PY{n}{keras}\PY{o}{.}\PY{n}{models}\PY{o}{.}\PY{n}{Sequential}\PY{p}{(}\PY{p}{[}
  \PY{n}{tf}\PY{o}{.}\PY{n}{keras}\PY{o}{.}\PY{n}{layers}\PY{o}{.}\PY{n}{Flatten}\PY{p}{(}\PY{n}{input\PYZus{}shape}\PY{o}{=}\PY{p}{(}\PY{l+m+mi}{28}\PY{p}{,} \PY{l+m+mi}{28}\PY{p}{)}\PY{p}{)}\PY{p}{,}
  \PY{n}{tf}\PY{o}{.}\PY{n}{keras}\PY{o}{.}\PY{n}{layers}\PY{o}{.}\PY{n}{Dense}\PY{p}{(}\PY{l+m+mi}{64}\PY{p}{,} \PY{n}{activation}\PY{o}{=}\PY{n}{tf}\PY{o}{.}\PY{n}{nn}\PY{o}{.}\PY{n}{relu}\PY{p}{)}\PY{p}{,}
  \PY{n}{tf}\PY{o}{.}\PY{n}{keras}\PY{o}{.}\PY{n}{layers}\PY{o}{.}\PY{n}{Dropout}\PY{p}{(}\PY{l+m+mf}{0.2}\PY{p}{)}\PY{p}{,}
  \PY{n}{tf}\PY{o}{.}\PY{n}{keras}\PY{o}{.}\PY{n}{layers}\PY{o}{.}\PY{n}{Dense}\PY{p}{(}\PY{l+m+mi}{4}\PY{p}{)}\PY{p}{,}
  \PY{n}{tf}\PY{o}{.}\PY{n}{keras}\PY{o}{.}\PY{n}{layers}\PY{o}{.}\PY{n}{Lambda}\PY{p}{(}\PY{n}{qsoftmax}\PY{p}{(}\PY{n}{basis4}\PY{p}{)}\PY{p}{)}
\PY{p}{]}\PY{p}{)}
\PY{n}{basis7\PYZus{}model} \PY{o}{=} \PY{n}{tf}\PY{o}{.}\PY{n}{keras}\PY{o}{.}\PY{n}{models}\PY{o}{.}\PY{n}{Sequential}\PY{p}{(}\PY{p}{[}
  \PY{n}{tf}\PY{o}{.}\PY{n}{keras}\PY{o}{.}\PY{n}{layers}\PY{o}{.}\PY{n}{Flatten}\PY{p}{(}\PY{n}{input\PYZus{}shape}\PY{o}{=}\PY{p}{(}\PY{l+m+mi}{28}\PY{p}{,} \PY{l+m+mi}{28}\PY{p}{)}\PY{p}{)}\PY{p}{,}
  \PY{n}{tf}\PY{o}{.}\PY{n}{keras}\PY{o}{.}\PY{n}{layers}\PY{o}{.}\PY{n}{Dense}\PY{p}{(}\PY{l+m+mi}{64}\PY{p}{,} \PY{n}{activation}\PY{o}{=}\PY{n}{tf}\PY{o}{.}\PY{n}{nn}\PY{o}{.}\PY{n}{relu}\PY{p}{)}\PY{p}{,}
  \PY{n}{tf}\PY{o}{.}\PY{n}{keras}\PY{o}{.}\PY{n}{layers}\PY{o}{.}\PY{n}{Dropout}\PY{p}{(}\PY{l+m+mf}{0.2}\PY{p}{)}\PY{p}{,}
  \PY{n}{tf}\PY{o}{.}\PY{n}{keras}\PY{o}{.}\PY{n}{layers}\PY{o}{.}\PY{n}{Dense}\PY{p}{(}\PY{l+m+mi}{7}\PY{p}{)}\PY{p}{,}
  \PY{n}{tf}\PY{o}{.}\PY{n}{keras}\PY{o}{.}\PY{n}{layers}\PY{o}{.}\PY{n}{Lambda}\PY{p}{(}\PY{n}{qsoftmax}\PY{p}{(}\PY{n}{basis7}\PY{p}{)}\PY{p}{)}
\PY{p}{]}\PY{p}{)}
\end{Verbatim}
\vspace{1mm}
Table \DUrole{ref}{tab:qoe} shows the mean of the accuracy over three training runs
of the validation data with training data in parentheses.
\begin{table}
  \begin{longtable*}{lcccc}
  \toprule
  \textbf{Number of} & \textbf{One Hot} & \textbf{7-Dimensional} & \textbf{4-Dimensional} \\
  \textbf{Epochs} & \textbf{Encoding} & \textbf{QOE} & \textbf{QOE} \\
  \midrule
  \endfirsthead
  10 & 97.53\% (97.30\%) & 97.24\% (96.94\%) & 95.65\% (95.15\%) \\
  20 & 97.68\% (98.02\%) & 97.49\% (97.75\%) & 95.94\% (96.15\%) \\
  \bottomrule
  \end{longtable*}
  \caption{Results of MNIST QOE Experiment \DUrole{label}{tab:qoe}}
\end{table}

From these results, it is clear that there is some degradation in performance as the number of dimensions is reduced, but clearly QOE can be used leading to a tradeoff between accuracy and resource reduction from the reduction of dimensionality.

\subsection{Extending to Spherical Encodings%
  \label{extending-to-spherical-encodings}%
}

\subsubsection{A Deeper Look at Softmax%
  \label{a-deeper-look-at-softmax}%
}

In principle, according to equation \DUrole{ref}{eq:argmax}, the dot product against a basis be it orthonormal or quasiorthonormal recovers the category from a potentially noisy encoded value. If one takes a deeper dive into equations \DUrole{ref}{eq:gsoftmax} and \DUrole{ref}{eq:qsoftmax}, it is interesting to see what these functions are doing. Figure \DUrole{ref}{fig:ortho} shows on the left, randomly selected values in a circle of radius 6. On the right shows the vectors after the softmax function is applied. Clearly with a few stragglers, most points either move very close to either of the basis vectors $(0,1)$ or $(1,0)$. For this graphic the radius of 6 is chosen because upon examining the input to the softmax function in the example above, the average magnitude of each component is 5.5. In reality, 6 is probably a conservatively low value. Due to the exponential term in the softmax function, even large components would make the separation more pronounced.\begin{figure}[]\noindent\makebox[\columnwidth][c]{\includegraphics[width=\columnwidth]{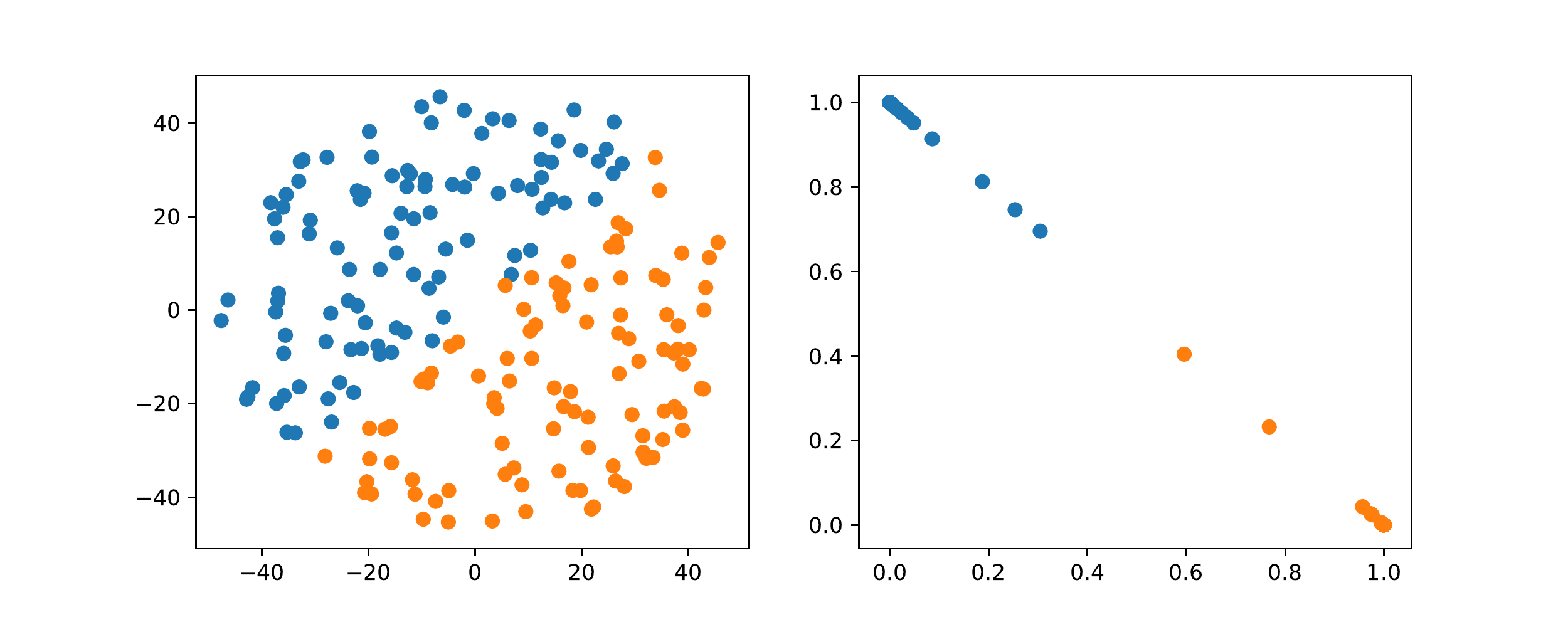}}
\caption{Softmax on an orthonormal basis \DUrole{label}{fig:ortho}}
\end{figure}

Similarly, figure \DUrole{ref}{fig:qortho} shows on the the same type of distribution of randomly selected values and the right shows the effect after a quasiorthonormal softmax is applied with three basis vectors. Since the qsoftmax function would map the two dimensional input into a three dimensional space, for graphics sake, the three dimensional vectors are mapped back down to two dimensions using the quasiorthonormal basis. Again with the exception of a few stragglers, most points move very close to one of the three basis vectors.\begin{figure}[]\noindent\makebox[\columnwidth][c]{\includegraphics[width=\columnwidth]{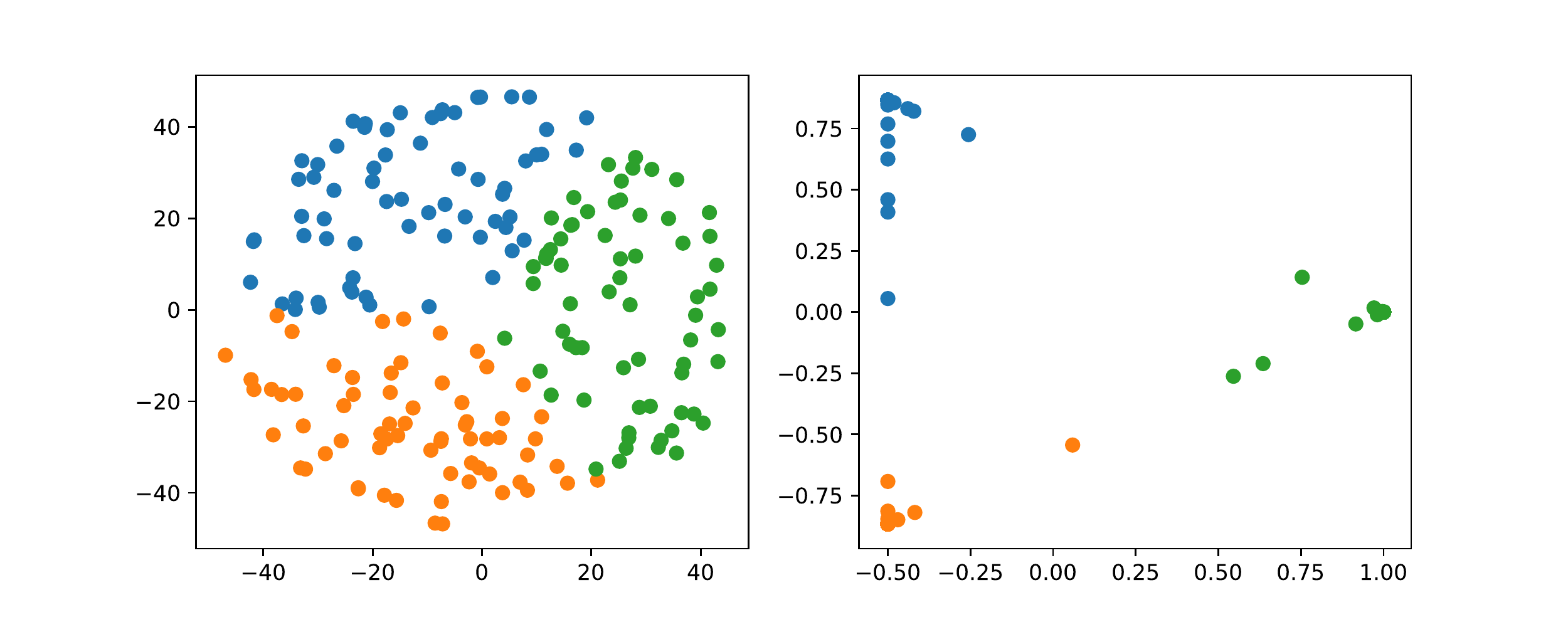}}
\caption{Softmax on a quasiorthogonal basis \DUrole{label}{fig:qortho}}
\end{figure}

It is clear from equation \DUrole{ref}{eq:gsoftmax} that the exponential term in the numerator  dominates which is why the softmax is effective.  But consider if $\mathbf{z}\cdot\mathbf{u}_i$ is much less than zero such as $\mathbf{z}=-5 \mathbf{u}_i$, then the exponential numerator for that term would severely attenuate the output even though the $\mathbf{z}$ lies along the same direction as $\mathbf{u}_i$.
Since the $\mathbf{z}\cdot\mathbf{u}_j=0$ for all other \emph{j}'s, it would still decode to $i$, but because the :math:\emph{i}-th term is so small, any noise could lead to decoding to a different value.

So why bring all this up? Clearly, values encode to negative values along a basis vector would prove problematic. But it offers the prospect of further reducing dimensionality by encoding to no just the basis a vector but also its antipodal vector. To further our graphical example, in figure \DUrole{ref}{fig:sphere}, we use $(1,0)$, $(0,1)$ and their antipodal vectors $(-1,0)$ and $(0,-1)$ to encode values and apply a softmax using those vectors.\begin{figure}[]\noindent\makebox[\columnwidth][c]{\includegraphics[width=\columnwidth]{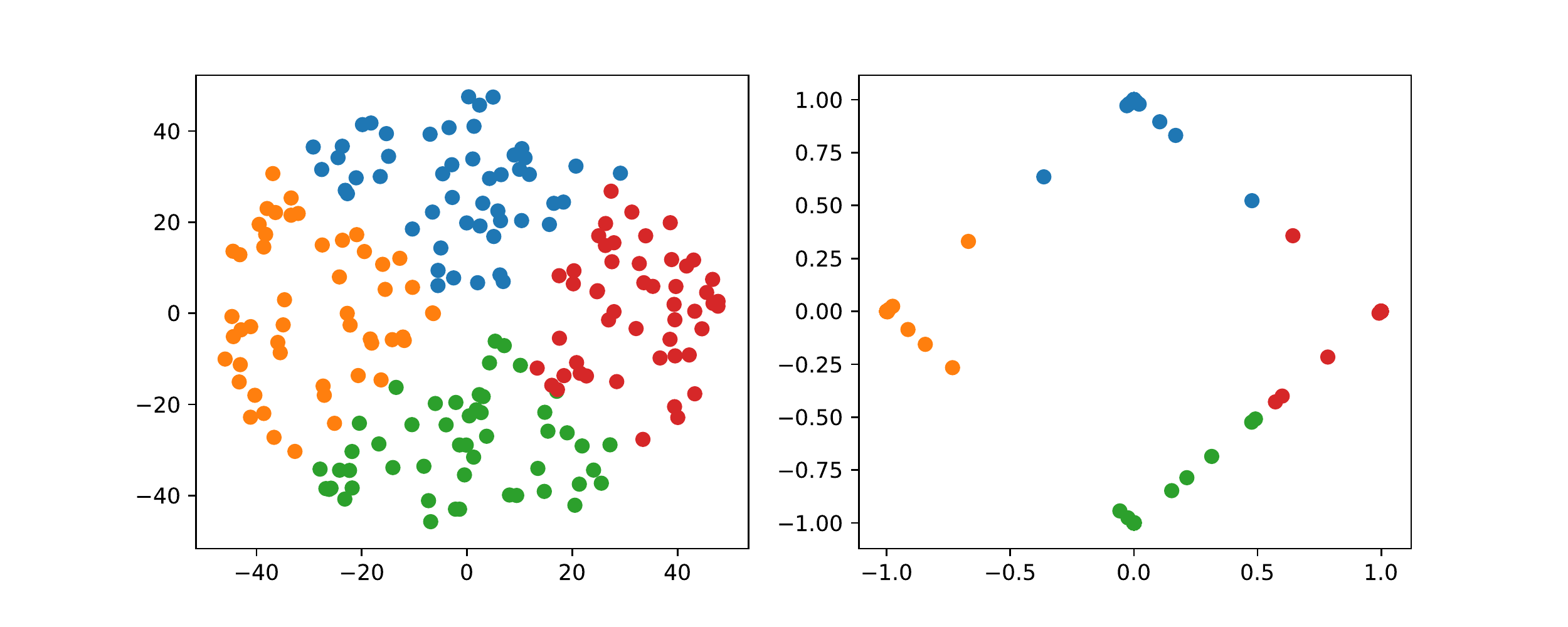}}
\caption{Softmax on encoded values using an orthonormal basis and antipodal points \DUrole{label}{fig:sphere}}
\end{figure}

Once again we see the power of the exponential term and most points move very close to one of the four encoding vectors.

Of course nothing comes for free, if a prediction gets confused
between two antipodal unit vectors, the result could be that they cancel
out and allow the noise to dictate the resulting category. By contrast,
for one-hot encoding, the result would get decoded as one of the two
possible values.

With this risk in mind, we can further extend the idea to a
quasiorthogonal basis by adding the antipodal vectors for each vector in
the basis. The result not only doubles the number of vectors that can be
used for encoding, it reduces the problem of finding a basis to that of
finding spherical codes.

\subsubsection{Spherical Codes%
  \label{spherical-codes}%
}

Spherical codes can be used in place of quasiorthonormal codes simply by allowing the $\mathbf{q}_i$ to be a collection of spherical codes not necessarily quasiorthonormal basis. Table \DUrole{ref}{tab:spherecar} shows how the example of the five Japanese car makes could be encoded with a simple spherical code.\begin{table}
\setlength{\DUtablewidth}{\tablewidth}
\begin{longtable*}[c]{p{0.145\DUtablewidth}p{0.168\DUtablewidth}p{0.203\DUtablewidth}}
\toprule
\textbf{%
Make} & \textbf{%
One-Hot} & \textbf{%
Spherical Code} \\
\midrule
\endfirsthead

Toyota & 
(1,0,0,0,0) & 
(1,0,0) \\

Honda & 
(0,1,0,0,0) & 
(-1,0,0) \\

Subaru & 
(0,0,1,0,0) & 
(0,1,0) \\

Nissan & 
(0,0,0,1,0) & 
(0,-1,0) \\

Mitsubishi & 
(0,0,0,0,1) & 
(0,0,1) \\
\bottomrule
\end{longtable*}
\caption{Examples of Spherical Codes \DUrole{label}{tab:spherecar}}\end{table}

Since spherical codes can substitute directly into the equations for QOE, it is a simple matter to implement spherical codes $\{\mathbf{s}_i\}$ instead of quasiorthonormal basis, $\{\mathbf{q}_i\}$. As such it is a simple matter to run the same experiment on the MNIST handwriting samples as we did for QOE. First, a set of codes are defined in an \texttt{ndarray} called \texttt{code5} and \texttt{code3}. The variable \texttt{code5} consists of the standard orthonormal basis in 5 dimensions along with their antipodal unit vector to produce a set of 10 vectors in 5 dimensions. The variable \texttt{code3} is taken from \DUrole{cite}{sphere} for the 3 dimensional spherical codes with 10 vectors. Once these codes are defined, they can be substituted for \texttt{basis4} and \texttt{basis7} in the sample code above. Table \DUrole{ref}{tab:spherecode} shows the results of the experiment with training accuracy shown in parentheses.
\begin{table}
  \begin{longtable*}{lcccc}
  \toprule
  \textbf{Number of} & \textbf{One Hot} & \textbf{5-Dimensional} & \textbf{3-Dimensional} \\
  \textbf{Epochs} & \textbf{Encoding} & \textbf{Spherical Code} & \textbf{Spherical Code} \\
  \midrule
  \endfirsthead
  10 & 97.53\% (97.30\%) & 96.51\% (96.26\%) & 95.37\% (94.83\%) \\
  20 & 97.68\% (98.02\%) & 96.82\% (97.11\%) & 95.74\% (95.83\%) \\
  \bottomrule
  \end{longtable*}
  \caption{Results of MNIST Spherical Coding Experiment \DUrole{label}{tab:spherecode}}
\end{table}

In this case, the 5-dimensional spherical codes performed close to the
one-hot encoding by not as closely as the 7-dimension QO codes. The
3-dimensional spherical codes performed on par with the 4-dimensional QO
codes.

While the extreme dimensionality reduction from 10 to 4 or 10 to 3 did
not yield comparable performance to one-hot encoding. More modest
reductions such as 10 to 7 and 10 to 5 did. It is worth considering that
quasiorthogonal or spherical codes are much harder to find in low
dimensions. One should note that, though we went from 10 to 7
dimensions, we did not fully exploit the space spanned by the
quasiorthogonal vector set. Otherwise, we would likely have had the
similar results if the categorical labels had a cardinality of 28 rather
than 10.

\subsection{Conclusion%
  \label{conclusion}%
}

These reduced dimensionality codes are not expected to improve accuracy when the training data is plentiful, but to save computation and representation by reducing the dimensionality of the coded category. As an example, in application such as autoencoders and specifically the imputation architectures presented by \DUrole{cite}{gondara} and \DUrole{cite}{lu} where the dimensionality not only dictates the number of outputs and inputs but also the number of hidden layers, a reduction in dimensionality has a profound impact on the size of the model used. Beyond that the reduced dimensionality codes such as QOE and spherical codes can address problems such as the curse of dimensionality and HDLSS where for small sample sizes it may improve accuracy.

Though for the exercises presented here, the reduction of dimensionality is modest and may not seem worth the trouble. The real benefit of these codes is in extremely high cardinality situations on the order of hundreds, thousands and beyond, such as zip codes, area codes, or medical diagnostic codes.

Practically speaking, while algorithms to generate spherical codes and quasiorthonormal sets are few, \DUrole{cite}{sphere} has a vast complication of spherical codes. At the extreme end, a spherical code with 196,560 vectors is available in 24 dimensions, enough to encode nearly 100,000 labels using QOE or 200,000 labels using spherical codes, \emph{in just 24 dimensions!}

In sum, QOE and spherical codes are useful tools to be included in a data scientists tool box along side other established categorical coding techniques.

Experiments and code samples are made available at \href{https://github.com/Westhealth/scipy2020/quasiorthonormal/}{https://github.com/Westhealth/scipy2020/quasiorthonormal}.
\bibliographystyle{alphaurl}
\bibliography{ourbib}

\end{document}